\documentclass[10pt, a4paper]{article}

\usepackage{lrec-coling2024} % this is the new style
\usepackage{latexsym}

\usepackage{placeins}
\usepackage{multirow}
\usepackage{graphicx}
\usepackage{pgf}
\usepackage{layouts}
\usepackage{mdframed}
\usepackage{lipsum}
\usepackage{paralist}
\usepackage{soul}

\usepackage[T1]{fontenc}
\usepackage[utf8]{inputenc}

\usepackage{microtype}

\usepackage{booktabs}
\usepackage{arydshln}
\usepackage{cleveref}
\crefformat{section}{\S#2#1#3} % see manual of cleveref, section 8.2.1
\crefformat{subsection}{\S#2#1#3}
\crefformat{subsubsection}{\S#2#1#3}

\usepackage{titlesec}
\titlespacing{\paragraph}{%
  0pt}{%              left margin
  0.5\baselineskip}{% space before (vertical)
  1em}%

\title{On Leveraging Encoder-only Pre-trained Language Models \\for Effective Keyphrase Generation}

\name{Di Wu, Wasi Uddin Ahmad, Kai-Wei Chang} 

\address{Department of Computer Science \\
University of California, Los Angeles \\ 
\{diwu, kwchang\}@cs.ucla.edu, wasiahmad@ucla.edu
}

\abstract{
This study addresses the application of encoder-only Pre-trained Language Models (PLMs) in keyphrase generation (KPG) amidst the broader availability of domain-tailored encoder-only models compared to encoder-decoder models. We investigate three core inquiries: (1) the efficacy of encoder-only PLMs in KPG, (2) optimal architectural decisions for employing encoder-only PLMs in KPG, and (3) a performance comparison between in-domain encoder-only and encoder-decoder PLMs across varied resource settings. Our findings, derived from extensive experimentation in two domains reveal that with encoder-only PLMs, although KPE with Conditional Random Fields slightly excels in identifying present keyphrases, the KPG formulation renders a broader spectrum of keyphrase predictions. Additionally, prefix-LM fine-tuning of encoder-only PLMs emerges as a strong and data-efficient strategy for KPG, outperforming general-domain seq2seq PLMs. We also identify a favorable parameter allocation towards model depth rather than width when employing encoder-decoder architectures initialized with encoder-only PLMs. The study sheds light on the potential of utilizing encoder-only PLMs for advancing KPG systems and provides a groundwork for future KPG methods. Our code and pre-trained checkpoints are released at \url{https://github.com/uclanlp/DeepKPG}.
\\ \newline }

\begin{document}

\setlength{\abovedisplayskip}{5pt}
\setlength{\belowdisplayskip}{5pt}
\maketitleabstract

\section{Introduction}
% Keyphrases are the phrases that condense salient information of a document. Because of their high information density, keyphrases have been widely used for indexing documents, linking to relevant information, or recommending products \citep{10.1145/1367497.1367723, 10.1145/1871437.1871754}. Keyphrases have also functioned as important features for information retrieval \citep{10.1145/312624.312671, 10.1145/1141753.1141800, kim-etal-2013-applying, Tang2017QALinkET, boudin-etal-2020-keyphrase}, text summarization \citep{10.5555/1039791.1039794}, document clustering \citep{Hammouda2005}, and text classification \citep{10.3115/1220175.1220243, wilson-etal-2005-recognizing, berend-2011-opinion}.

Keyphrases are phrases that condense salient information of a document. Given their capability to capture rich information, researchers have adopted keyphrases in various applications such as document indexing, information linking, and recommendation systems \citep{10.1145/1367497.1367723, 10.1145/1871437.1871754}. Additionally, keyphrases have manifested as important attributes in information retrieval \citep{10.1145/312624.312671, 10.1145/1141753.1141800, kim-etal-2013-applying, Tang2017QALinkET, boudin-etal-2020-keyphrase}, text summarization \citep{10.5555/1039791.1039794}, document clustering \citep{Hammouda2005}, and text categorization \citep{10.3115/1220175.1220243, wilson-etal-2005-recognizing, berend-2011-opinion}.

% In practice, keyphrases are often divided into two types. A keyphrase is a \textit{present keyphrase} if it appears as a span in the document, or an \textit{absent keyphrase} otherwise (Figure \ref{example-case}). The task \textit{keyphrase extraction} (KPE) requires a model to extract present keyphrases. \citet{meng-etal-2017-deep} introduce \textit{keyphrase generation} (KPG), which requires predicting all the present and absent keyphrases.

% \input{figures/example_kpgen_case}

Traditionally, keyphrases are classified into two categories. A \textit{present keyphrase} is one that is directly extractable from the document, while an \textit{absent keyphrase} does not have a direct representation within the text (Figure \ref{example-case}). The task of \textit{keyphrase extraction} (KPE) requires models to identify present keyphrases. \citet{meng-etal-2017-deep} extended this by introducing \textit{keyphrase generation} (KPG), aiming to predict both present and absent keyphrases.

\begin{figure}[!t]
    \centering
    \small
    \vspace{2mm}
    \resizebox{\linewidth}{!}{
    \begin{tabular}{ p{0.98\linewidth} }
    \toprule \\
    [-2ex]
    \textbf{Document title} \\
    \hdashline \\
    [-2ex]
    J.F.K. Workers Moved Drugs, Authorities Say \\
    \midrule
    \textbf{Document body} \\
    \hdashline \\
    [-2ex]
    Airline employees exploited weaknesses in security procedures to help a New York drug ring \textcolor{blue}{smuggle} \textcolor{blue}{heroin} and cocaine through \textcolor{blue}{Kennedy International Airport}, federal authorities charged yesterday. ... \\
    \midrule \\ [-2ex]
    \textbf{Present and Absent Keyphrases} \\
    \hdashline \\
    [-2ex]
    \textcolor{blue}{smuggling}, \textcolor{blue}{heroin}, \textcolor{blue}{kennedy international airport}, \textcolor{red}{drug abuse and traffic}, \textcolor{red}{crime and criminals}, \textcolor{red}{cocaine and crack cocaine} \\
    \bottomrule
    \end{tabular}}
    \vspace{-4mm}
    \caption{
    An example of a news article with its present and absent keyphrases highlighted in blue and red, respectively.
    }
    \vspace{-2mm}
    \label{example-case}
    \vspace{-2mm}
\end{figure}

% Since the advent of pre-trained language models (PLM), the KPE and KPG community has quickly embraced them for unsupervised KPE \citep{8954611,liang-etal-2021-unsupervised}, KPE via sequence labeling \citep{arxiv.1910.08840,9481005}, and KPG via sequence-to-sequence (seq2seq) generation \citep{9443960,arxiv.2004.10462,2201.05302}. In particular, PLM-based KPG methods have exhibited promising performance on zero-shot \citep{kulkarni-etal-2022-learning}, multilingual \citep{gao-etal-2022-retrieval}, and low-resource \citep{wu-etal-2022-representation} KPG.
In recent years, the emergence of pre-trained language models (PLMs) has revolutionized the KPE and KPG landscapes, leading to their application in unsupervised KPE \citep{8954611,liang-etal-2021-unsupervised}, sequence-labeled KPE \citep{arxiv.1910.08840,9481005}, and sequence-to-sequence KPG \citep{9443960,arxiv.2004.10462,2201.05302}. Remarkably, PLM-based KPG methodologies have exhibited promising performance, especially in scenarios such as zero-shot \citep{kulkarni-etal-2022-learning}, multilingual \citep{gao-etal-2022-retrieval}, low-resource \citep{wu-etal-2022-representation}, and cross-domain \citep{meng-etal-2023-general} settings.

% Nevertheless, real world applications of PLM-based keyphrase systems often face a dilemma. On one hand, KPG models are highly valued due to their ability to generate absent keyphrases, which are observed to bear a high utility in retrieval systems \citep{boudin-gallina-2021-redefining}. On the other hand, the availability of domain-specific PLMs \citep{gururangan-etal-2020-dont} strongly prefers KPE methods, which often employ encoder-only PLMs like BERT \citep{devlin-etal-2019-bert}, over KPG methods, which mainly use prefix-LMs \citep{wu-etal-2021-unikeyphrase} or sequence-to-sequence PLMs \citep{zhao-etal-2022-keyphrase}: a large amount of encoder-only models are available in a fine-grained set of domains (Figure \ref{various-bert-models}), while domain-specific encoder-decoder models are much rarer.

However, the deployment of PLM-driven systems for keyphrase prediction in practical applications presents a dilemma. While the ability of KPG models to generate absent keyphrases is lauded, particularly due to its impact in retrieval systems \citep{boudin-gallina-2021-redefining}, there is a nuanced inclination towards KPE techniques for domain-specific applications. Such techniques predominantly harness in-domain encoder-only PLMs like BERT \citep{devlin-etal-2019-bert, gururangan-etal-2020-dont}, as opposed to the KPG methods which often employ prefix-LMs \citep{wu-etal-2021-unikeyphrase} or sequence-to-sequence PLMs \citep{zhao-etal-2022-keyphrase}. The abundance of encoder-only models across specific domains accentuates this preference (Figure \ref{various-bert-models}), while domain-oriented encoder-decoder models remain relatively scarce.

Motivate by this challenge, this paper studies the following research questions:

% \begin{itemize}
% \item The feasibility of deploying encoder-only PLMs for KPG and assessing their performance in extracting present keyphrases compared to KPE models.
% \item The optimal architectural configuration for implementing encoder-only PLMs within KPG.
% \item A comparative analysis between in-domain encoder-only models and encoder-decoder PLMs in both abundant and resource-limited environments.
% \end{itemize}

% \begin{compactitem}
\begin{itemize}
    \item Can we use encoder-only PLMs for KPG and achieve a similar present keyphrase performance as using them for KPE? 
    \item What is the best architectural choice for using encoder-only PLMs for KPG?
    \item How do in-domain encoder-only PLMs compare to encoder-decoder PLMs in rich and low resource settings?
\end{itemize}
% \end{compactitem}

To answer these questions, we investigate four formulations for KPE and KPG with encoder-only PLMs. For KPE, we fine-tune on sequence labeling with or without Conditional Random Field \citep{10.5555/645530.655813}. For KPG, we fine-tune either using prefix-LM style attention masks \citep{arxiv.1905.03197}, or initialize an encoder-decoder architecture with encoder-only PLMs \citep{rothe-etal-2020-leveraging}. With extensive experiments on datasets covering the science and news domain, our main findings are summarized as follows.

% \begin{compactenum}
\begin{enumerate}
\item For present keyphrases, KPE with Conditional Random Fields (CRF) exhibits a slight advantage over KPG in terms of macro F1@M. However, training encoder-only models for KPG produce much more keyphrase predictions including both present and absent keyphrases. 
\item Prefix-LM fine-tuning of encoder-only PLMs is a strong and data-efficient KPG method, even outperforming general-domain seq2seq PLMs of the same size. 
\item For encoder-decoder architecture initialized with encoder-only PLMs, how parameters are allocated strongly affects the performance. Specifically, model \textit{depth} should be prioritized over the width, and a \textit{deep encoder with a shallow decoder} outperforms the reverse for keyphrase quality and inference latency.
\item We conducted an evaluation of applying the encoder-only SciBERT and NewsBERT for KPG inside and outside of their pre-training domain. Compared to the general-domain BERT, both models demonstrate better in-domain performance. However, SciBERT transfers well to the news domain while NewsBERT cannot transfer to the science domain. 
\end{enumerate}
% \end{compactenum}

We hope this empirical study can shed light on the opportunities of leveraging encoder-only PLMs for building KPG systems, facilitating the development of more effective approaches. Our code and the pre-trained NewBERT model will be released at \url{https://github.com/uclanlp/DeepKPG}.

% \begin{figure}[h]
%     \centering
%     \small
%     \resizebox{\linewidth}{!}{
%     \begin{tabular}{ p{0.98\linewidth} }
%     \toprule
%     \textbf{Academic domain} \\
%     \hdashline \\
%     [-2ex]
%     SciBERT \citep{beltagy-etal-2019-scibert}, BioBERT \citep{Lee_2019}, ChemBERTa \citep{2010.09885}, [Bio|CS]\_RoBERTa \citep{gururangan-etal-2020-dont}, BioMegatron \citep{shin-etal-2020-biomegatron}, MatSciBERT \citep{2109.15290}, PubMedBERT \citep{Gu_2022}, MatBERT \citep{TREWARTHA2022100488}, BatteryBERT \citep{huang2022batterybert}  \\
%     \midrule \\ 
%     [-2ex]
%     \textbf{Social domain}  \\
%     \hdashline \\
%     [-2ex]
%     ClinicalBERT \citep{alsentzer-etal-2019-publicly}, FinBERT \citep{2006.08097}, LEGAL-BERT \citep{chalkidis-etal-2020-legal}, JobBERT \citep{zhang-etal-2022-skillspan}, PrivBERT \citep{srinath-etal-2021-privacy}, SportsBERT \citep{sportsbert} \\
%     % \footnote{\textcolor{red}{(https://huggingface.co/microsoft/SportsBERT)}} \\
%     \midrule \\ 
%     [-2ex] 
%     \textbf{Web domain}  \\
%     \hdashline \\
%     [-2ex]
%     Twitter-roberta \citep{barbieri-etal-2020-tweeteval}, BERTweet \citep{nguyen-etal-2020-bertweet},  [News|Reviews]\_RoBERTa \citep{gururangan-etal-2020-dont}, HateBERT \citep{caselli-etal-2021-hatebert} \\
%     \bottomrule
%     \end{tabular}}
%     \vspace{-2mm}
%     \caption{
%     Domain-specific encoder-only PLMs are available in a variety of domains. No prior work considered using these "domain experts" for keyphrase generation.
%     }
%     % \vspace{-2mm}
%     \label{various-bert-models}
%     % \vspace{-2mm}
% \end{figure}

\begin{figure}[t]
    \centering
    % \small
    \resizebox{1.0\linewidth}{!}{
    \begin{tabular}{ l | l | l  }
    \toprule
    \textbf{Model Name} & \textbf{Reference} &\textbf{Domain} \\ 
    \midrule
    SciBERT & \citet{beltagy-etal-2019-scibert} & Science \\
    BioBERT & \citet{Lee_2019} & Science \\
    ChemBERTa & \citet{2010.09885} & Science \\
    Bio\_RoBERTa & \citet{gururangan-etal-2020-dont} & Science \\
    CS\_RoBERTa & \citet{gururangan-etal-2020-dont} & Science \\
    BioMegatron & \citet{shin-etal-2020-biomegatron} & Science \\
    MatSciBERT & \citet{2109.15290} & Science \\
    PubMedBERT & \citet{Gu_2022} & Science \\
    MatBERT & \citet{TREWARTHA2022100488} & Science \\ 
    BatteryBERT & \citet{huang2022batterybert} & Science \\
    \hdashline
    ClinicalBERT & \citet{alsentzer-etal-2019-publicly} & Social \\
    FinBERT & \citet{2006.08097} & Social  \\
    LEGAL-BERT & \citet{chalkidis-etal-2020-legal} & Social  \\
    JobBERT & \citet{zhang-etal-2022-skillspan} & Social  \\
    PrivBERT & \citet{srinath-etal-2021-privacy} & Social  \\
    SportsBERT & \citet{sportsbert} & Social  \\
    \hdashline
    Twitter-roberta & \citet{barbieri-etal-2020-tweeteval} & Web \\
    BERTweet & \citet{nguyen-etal-2020-bertweet} & Web  \\
    News\_RoBERTa & \citet{gururangan-etal-2020-dont} & Web  \\
    NewsBERT & this work & Web \\
    Reviews\_RoBERTa & \citet{gururangan-etal-2020-dont} & Web  \\
    HateBERT & \citet{caselli-etal-2021-hatebert} & Web  \\
    \bottomrule
    \end{tabular}}
    % \vspace{-2mm}
    \caption{
    Domain-specific encoder-only PLMs are available in a variety of domains. No prior work considered using these "domain experts" for KPG.  In this paper, we show that these specialized encoder-only PLMs can be used to build strong and resource-efficient KPG models.
    }
    % \vspace{-2mm}
    \label{various-bert-models}
    % \vspace{-2mm}
\end{figure}

\section{Related Work}

% \diwu{Need to discuss more how this work position in the current literature. }

\paragraph{Keyphrase Extraction}
Early work on KPE mainly followed a pipelined approach. First, keyphrase candidates (usually noun phrases) are extracted by rules such as regular expression matching on part-of-speech tags. Then, various scoring methods are used to rank the candidates. The ones with the highest scores are returned as keyphrase predictions \citep{hulth-2003-improved,mihalcea-tarau-2004-textrank,10.5555/1620163.1620205,bougouin-etal-2013-topicrank,8954611,boudin-2018-unsupervised, liang-etal-2021-unsupervised}. In terms of supervised approaches, one line of research adopts the sequence labeling formulation, removing the need for selecting candidates \citep{zhang-etal-2016-keyphrase,luan-etal-2017-scientific,arxiv.1910.08840}. Others instead aim to learn a more capable ranking function with pre-trained language models \citep{song-etal-2021-importance, song-etal-2022-hyperbolic}.

\paragraph{Keyphrase Generation}

\citet{meng-etal-2017-deep} propose the task of Deep Keyphrase Generation and a strong baseline model CopyRNN. Later works improve the architecture by adding correlation constraints \citep{chen-etal-2018-keyphrase} and linguistic constraints \citep{zhao-zhang-2019-incorporating}, exploiting learning signals from titles \citep{ye-wang-2018-semi, Chen2019TitleGuidedEF}, and hierarchical modeling the phrases and words \citep{chen-etal-2020-exclusive}. \citet{yuan-etal-2020-one} reformulate the problem as generating a sequence of keyphrases, while \citet{ye-etal-2021-one2set} further introduces a set generation reformulation to remove the influence of difference target phrase ordering. Other works include incorporating reinforcement learning \citep{chan-etal-2019-neural, luo-etal-2021-keyphrase-generation}, GANs \citep{swaminathan-etal-2020-preliminary}, and unifying KPE with KPG \citep{chen-etal-2019-integrated,ahmad-etal-2021-select, wu-etal-2021-unikeyphrase}. \citet{meng-etal-2021-empirical} conduct an empirical study on architecture, generalizability, phrase order, and decoding strategies, mainly focusing on RNNs and Transformers trained from scratch.

\paragraph{KPE and KPG with PLMs} More recently, \citet{arxiv.1910.08840}, \citet{9443960}, \citet{arxiv.2004.10462}, and \citet{9481005} have considered using pre-trained BERT \citep{devlin-etal-2019-bert} for keyphrase extraction and generation. \citet{wu-etal-2021-unikeyphrase} propose to fine-tune a prefix-LM to jointly extract and generate keyphrases. In addition, \citet{2201.05302}, \citet{kulkarni-etal-2022-learning}, \citet{wu-etal-2022-representation}, and \citet{gao-etal-2022-retrieval} use seq2seq PLMs such as BART or T5 in their approach. \citet{kulkarni-etal-2022-learning} use keyphrase generation as a pre-training task to learn strong PLM-based representations. \citet{wu-etal-2023-rethinking} investigate why and how encoder-decoder PLMs can be effective for KPG, focusing on model selection and decoding strategies. In this paper, we aim to bridge the gap in understanding how encoder-only PLMs could be best utilized for KPG.
\section{Evaluation Setup}
In this section, we formulate the KPE and KPG tasks and introduce the evaluation setup. We end by discussing the PLMs we are considering.

% In this section, we first formulate KPE and KPG in \cref{section-task-definition} and then introduce the considered PLM-based approaches in \cref{section-kpe-methods} and \cref{section-kpg-methods}.

\subsection{Problem Definition}
\label{section-task-definition}
We view a keyphrase example as a triple \textbf{$(\mathbf{x},\mathbf{y_p},\mathbf{y_a})$}, corresponding to the input document $\mathbf{x}=(x_1,x_2,...,x_d)$, the set of present keyphrases $\mathbf{y_p}=\{y_{p_1}, y_{p_2}, ..., y_{p_m}\}$, and the set of absent keyphrases $\mathbf{y_a}=\{y_{a_1}, y_{a_2}, ..., y_{a_n}\}$. For both keyphrase extraction and generation, $\mathbf{x}$ consists of the title and the document body, concatenated with a special $[sep]$ token. Following \citet{meng-etal-2017-deep}, each $y_{p_i}\in\mathbf{y_p}$ is a subsequence of $\mathbf{x}$, and each $y_{a_i}\in\mathbf{y_a}$ is not a subsequence of $\mathbf{x}$.

Using this formulation, the \textbf{keyphrase extraction} (KPE) task requires the model to predict $\mathbf{y_p}$. On the other hand, the \textbf{keyphrase generation} (KPG) task requires the prediction of $\mathbf{y_p} \cup \mathbf{y_a}$. 

\subsection{Benchmarks}
We evaluate using two widely-used KPG benchmarks in the science and the news domains. Table \ref{tab:test-sets-statistics} summarizes the statistics of the testing datasets.

\paragraph{\textbf{SciKP}} \citet{meng-etal-2017-deep} introduce KP20k, which contains 500k Computer Science papers. Following their work, we train on KP20k and evaluate on the KP20k test set.

% as well as four out-of-distribution testing datasets: Inspec \citep{10.3115/1119355.1119383}, Krapivin \citep{Krapivin2009LargeDF}, NUS \citep{10.1007/978-3-540-77094-7_41}, and SemEval \citep{kim-etal-2010-semeval}.

\paragraph{\textbf{KPTimes}} Introduced by \citet{gallina-etal-2019-kptimes}, KPTimes is a keyphrase dataset in the news domain containing over 250k examples. We train on the KPTimes train set and report the performance on the union of the KPTimes test set and the out-of-distribution test set JPTimes.

% \paragraph{\textbf{StackEx}} We also establish the performance of various PLMs on the StackEx dataset \citep{yuan-etal-2020-one}. The results are presented in the appendix.

\setlength{\tabcolsep}{3pt}
\begin{table}[t]
    \centering
    % \small
    % \resizebox{\linewidth}{!}
    {%
    \begin{tabular}{l | c  c  c c}
    \hline
    Dataset & \#Examples & \#KP & $\%$AKP & |KP|  \\
    \hline
    KP20k & 20000 & 5.3 & 37.1 & 2.0  \\
    % Inspec & 500 & 9.8 & 26.4 & 2.5 \\
    % Krapivin & 400 & 5.9 & 44.3 & 2.2\\
    % NUS & 211 & 11.7 & 45.6 & 2.2  \\
    % SemEval & 100 & 14.7 & 57.4 & 2.4 \\
    % \hdashline
    KPTimes & 20000 & 5.0 & 37.8 & 2.0 \\
    % \hdashline
    % StackEx & 16000 & 2.7 & 42.5 & 1.3 \\
    \hline
    \end{tabular}
    }
    % \vspace{-2mm}
    \caption{Test sets statistics. \#KP, $\%$AKP, and |KP| refers to the average number of keyphrases per document, the percentage of absent keyphrases, and the average number of words in each keyphrase.}
    \label{tab:test-sets-statistics}
    % \vspace{-2mm}
\end{table}

\subsection{Evaluation}
Each method's predictions are normalized into a sequence of present and absent keyphrases. The phrases are ordered by the position in the source document for the sequence labeling approaches to obtain the keyphrase predictions. Then, we apply the Porter Stemmer \citep{Porter1980AnAF} to the output and target phrases and remove the duplicated phrases from the output. Following \citet{chan-etal-2019-neural} and \citet{chen-etal-2020-exclusive}, we separately report the macro-averaged F1@5 and F1@M scores for present and absent keyphrases. For all the results except the ablation studies, we train with three different random seeds and report the averaged scores. 

\subsection{Baselines}
We consider four supervised KPG baselines.

\textbf{CatSeq} \citep{yuan-etal-2020-one} is a CopyRNN \citep{meng-etal-2017-deep} trained on generating keyphrases as a sequence, separated by the separator token.

\textbf{ExHiRD-h} \citep{chen-etal-2020-exclusive} improves CatSeq with a hierarchical decoding framework and a hard exclusion mechanism to reduce duplicates.

\textbf{Transformer} \citep{ye-etal-2021-one2set} is the self-attention based seq2seq model \citep{vaswani2017attention} with copy mechanism \citep{see-etal-2017-get}. 

\textbf{SetTrans} \citep{ye-etal-2021-one2set} performs order-agnostic KPG. The model uses control codes trained via a k-step target assignment algorithm.

For KPE, we also include a range of ranking-based unsupervised KPE methods and a \textbf{Transformer} as baseline references. As the main focus of the paper is studying encoder-only PLMs, we only compare with the strongest non-PLM baselines. % in the main text and present the other baselines in the appendix. Appendix section \ref{appendix_baselines} introduces all the baselines and their implementation details. 

\subsection{Considered PLMs}
\label{section-plms}

\paragraph{Encoder-Only PLMs} We study \textbf{BERT} \citep{devlin-etal-2019-bert}, \textbf{SciBERT} \citep{beltagy-etal-2019-scibert}, \textbf{NewsBERT} (details below), and five pre-trained BERT checkpoints from \citet{arxiv.1908.08962} with the hidden size 768, 12 attention heads per layer, and 2, 4, 6, 8, and 10 layers. 

\paragraph{Encoder-Decoder PLMs} As this paper focuses on encoder-only PLMs, we only include \textbf{BART} \citep{lewis-etal-2020-bart} and its in-domain variations \textbf{SciBART} \citep{wu-etal-2023-rethinking} and \textbf{NewsBART} (details below) as references. 

\paragraph{Domain-Specific PLMs} Inspired by \citet{wu-etal-2023-rethinking}, we pre-train \textbf{NewsBART} using the RealNews dataset \citep{zellers2019neuralfakenews}, which contains around 130GB of news text from 2016 to 2019. We follow \citet{wu-etal-2023-rethinking} for data cleaning and pre-processing. For training, we mask 30\% of tokens and sample the spans from a Poisson distribution ($\lambda=3.5$). For 10\% of the masking spans selected to mask, we replace them with a random token instead of the special \texttt{<mask>} token. We set the maximum sequence length to 512. Starting from BART-base, the model is trained for 250k steps using the Adam optimizer with batch size 2048, learning rate 3e-4, 10k warm-up steps, and polynomial learning rate decay. The pre-training was conducted on 8 Nvidia A100 GPUs and costed approximately 5 days.

We also pre-train a \textbf{NewsBERT} model using the same data on masked language modeling with 15\% dynamic masking. Starting from BERT-base \citep{devlin-etal-2019-bert}, the model is trained for 250k steps with batch size 512, learning rate 1e-4, 5k warm-up steps, and linear learning rate decay. We use the Adam optimizer for pre-training. The pre-training was conducted on 8 Nvidia V100 GPUs and costed approximately 7 days.

\section{Modeling Approaches}
Next, we introduce the modeling methods we adopt in this paper for KPE and KPG. 

\subsection{Keyphrase Extraction (KPE)}
\label{section-kpe-methods}
For KPE, we use a sequence labeling formulation. Each $x_i\in \mathbf{x}$ is assigned a label $c_i\in\{B_{kp}, I_{kp}, O\}$ depending on $x$ being the beginning token of a present keyphrase, the subsequent token of a present keyphrase, or otherwise. The model is required to predict the label for each token.

In this paper, we fine-tune three encoder-only PLMs: \textbf{BERT} \citep{devlin-etal-2019-bert}, \textbf{SciBERT} \citep{beltagy-etal-2019-scibert}, and \textbf{NewsBERT} (section \ref{section-plms}) \footnote{In this study, we use the base variants of all the encoder-only models unless otherwise specified.}. We add a fully connected layer for each model to project the hidden representation of every token into three logits representing $B_{kp}$, $I_{kp}$, and $O$. The model is trained on the cross-entropy loss. We also experiment with using Conditional Random Field \citep{10.5555/645530.655813} to model the sequence-level transitions. We use \textbf{+CRF} to refer to this setting.

\subsection{Keyphrase Generation (KPG)}
\label{section-kpg-methods}

Following \citet{yuan-etal-2020-one}, we use a special separator token \texttt{;} to join all the keyphrases in a sequence $\mathbf{y}=(y_{p_1}$ \texttt{;} $...$  \texttt{;} $y_{p_m}$ \texttt{;} $y_{a_1}$ \texttt{;} $...$  \texttt{;} $y_{a_m})$. Using this sequence generation formulation, we fine-tune the encoder-decoder PLMs for KPG. The models are trained with the cross-entropy loss for generating the target sequence $\mathbf{y}$. Next, we discuss the methods we explore for training encoder-only PLMs on KPG. 

\paragraph{BERT2BERT}
We construct seq2seq keyphrase generation models by separately initializing the encoder and the decoder with encoder-only PLMs. Following \citet{rothe-etal-2020-leveraging}, we add cross-attention layers to the decoder. We use five pre-trained BERT checkpoints from \citet{arxiv.1908.08962} with hidden size 768, 12 attention heads per layer, and 2, 4, 6, 8, and 10 layers. \textbf{B2B-$e$+$d$} denotes a BERT2BERT model with an $e$-layer pre-trained BERT as the encoder and a $d$-layer pre-trained BERT as the decoder. We use BERT2RND (\textbf{B2R}) to denote randomly initializing the decoder and RND2BERT (\textbf{R2B}) to denote randomly initializing the encoder. The models are fine-tuned using the same sequence generation formulation. 

\setlength{\tabcolsep}{1pt}
\begin{table*}[h]
    \small
    \centering
    \setlength{\tabcolsep}{2pt}
    \resizebox{0.75\linewidth}{!} {%
    \begin{tabular}{l | c | c c c | c c c }
    \hline
    \multirow{2}{*}{Method} &
    \multirow{2}{*}{|M|} & \multicolumn{3}{c|}{\textbf{KP20k}} & \multicolumn{3}{c}{\textbf{KPTimes}} \\
    % \cmidrule{3-4} \cmidrule{5-6}
    & & \#KP & F1@5 & F1@M & \#KP &  F1@5 & F1@M \\
    % \midrule
    \hline
    \multicolumn{6}{l}{\textbf{Keyphrase extraction}} \\ \hline
    % Transformer & 110M & 23.5 & 33.8 & 28.8 & 42.7 \\
    BERT & 110M & (3.4, 0.0) & 27.9 & 38.9 & (2.5, 0.0) & 34.0 & 49.3\\
    % RoBERTa & 125M & 27.9 & 39.4 & 33.2 & 48.9 \\
    SciBERT & 110M & (3.1, 0.0) & 28.6 & 40.5 & (2.2, 0.0) & 31.8 & 47.7 \\
    NewsBERT & 110M & (3.0, 0.0) & 25.8 & 37.5 & (2.5, 0.0) & 34.5 & 50.4 \\
    \hdashline
    % Transformer+CRF & 110M & 24.9 & 36.4 & 28.2 & 43.2 \\
    BERT+CRF & 110M & (3.2, 0.0) & 28.0 & 40.6 & (2.4, 0.0) & 33.9 & 49.9  \\
    % RoBERTa+CRF & 125M & 26.7 & 39.0 & 32.4 & 48.4 \\
    SciBERT+CRF & 110M & (2.8, 0.0) & 28.4 & \textbf{42.1} & (2.1, 0.0) & 31.8 & 48.1\\
    NewsBERT+CRF & 110M & (2.8, 0.0) & 26.8 & 39.7 & (2.4, 0.0) & \textbf{34.9} & \textbf{50.8} \\
    \hline
    \multicolumn{6}{l}{\textbf{Present keyphrase generation}} \\ \hline
    BERT-G & 110M & (4.1, 0.9) & 31.3 & 37.9 & (2.3, 2.2) & 32.3 & 47.4 \\
    % RoBERTa-G & 125M & 28.8 & 36.9 & 33.0 & 48.2 \\
    SciBERT-G & 110M & (4.3, 1.1) & \textbf{32.8} & 39.7 & (2.4, 2.0) & 33.0 & 48.4 \\ 
    NewsBERT-G & 110M & (4.2, 0.7) & 29.9 & 36.8 & (2.4, 2.1) & 33.0 & 48.0 \\   
    \hline
    \end{tabular}
    }
    \caption{
    Present keyphrase performance of encoder-only PLM-based sequence labeling and sequence generation approaches. \#KP = average number of (present, absent) keyphrase predictions per document. The best results are boldfaced. We find that the prefix-LM approach allows the encoder-only models to generate much more keyphrases without greatly sacrificing F1@M.}
    \label{tab:main-kpe-results}
    % \vspace{-2mm}
\end{table*}

\paragraph{Prefix-LM}
\citet{arxiv.1905.03197} propose jointly pre-training for unidirectional, bidirectional, and seq2seq language modeling by controlling attention mask patterns. In the seq2seq setup, the input is $\mathbf{x}\ [eos]\ \mathbf{y}$. The attention mask is designed such that tokens in $\mathbf{x}$ are only allowed to attend to $\mathbf{x}$, and that tokens in $\mathbf{y}$ are allowed to attend to tokens on their left. Using this formulation, we fine-tune encoder-only PLMs for seq2seq keyphrase generation. Following \citet{arxiv.1905.03197}, we mask and randomly replace tokens from $\mathbf{y}$ and train the model on the cross-entropy loss between its reconstruction and the original sequence. We call our models \textbf{BERT-G}, \textbf{SciBERT-G}, and \textbf{NewsBERT-G}. Our approach is different from \citet{wu-etal-2021-unikeyphrase} in that (1) we use encoder-only PLMs that are not pre-trained on prefix-LM and (2) we consider letting the model directly generate the present keyphrases instead of adding new layers dedicated to KPE. 

% For a more comprehensive comparison, we also experiment on the UniLMv2 model without the relative position bias \citep{arxiv.2002.12804}, denoted as \textbf{UniLM}. 

\subsection{Implementation Details}
\paragraph{Keyphrase Extraction} We implemented our models with Huggingface Transformers\footnote{\url{https://github.com/huggingface/transformers}} and TorchCRF\footnote{\url{https://github.com/s14t284/TorchCRF}}. The models are trained for ten epochs with early stopping. We use a learning rate of 1e-5 with linear decay and batch size 32 for most models (see appendix for all the hyperparameters). We use AdamW with $\beta_1=0.9$ and $\beta_2=0.999$.

\paragraph{Keyphrase Generation}
For BART and BERT2BERT, we use Huggingface Transformers to implement the fine-tuning. To fine-tune SciBART-base, SciBART-large, and NewsBART-base, we use the fairseq\footnote{\url{https://github.com/facebookresearch/fairseq}}. For BERT-G, SciBERT-G, NewsBART-G, and UniLM, we implement the training based on \citet{arxiv.1905.03197}'s original implementations \footnote{\url{https://github.com/microsoft/unilm}}. We set the maximum source and target length to 464 tokens and 48 tokens, respectively. We mask 80\% of the target tokens and randomly replace an additional 10\%. We use the AdamW optimizer.

\paragraph{Hyperparameters} For each of the PLM-based KPE and KPG methods, we perform a careful hyperparameter search over the learning rate, learning rate schedule, batch size, and warm-up steps. The corresponding search spaces are \{1e-5, 5e-4\}, \{linear, polynomial\}, \{16, 32, 64, 128\}, and \{500, 1000, 2000, 4000\}. For all models, we use the Adaom optimizer, apply a linear learning rate schedule, and do early stopping. The best hyperparameters found are presented in Table \ref{tab:hyperparams}. 

The fine-tuning experiments are run on a local GPU server with Nvidia GTX 1080 Ti and RTX 2080 Ti GPUs. We use at most 4 GPUs and gradient accumulation to achieve the desired batch sizes. We use greedy decoding for all the models. For evaluation, we follow \citet{chan-etal-2019-neural}'s implementation.

\section{Results}
We aim to address the following questions.
\begin{enumerate}
    \item Is the KPG formulation less suitable to encoder-only PLMs compared to KPE?
    \item Can encoder-only PLMs generate better keyphrases than encoder-decoder PLMs?
    \item What is the best parameter allocation strategy for using encoder-decoder PLMs to balance the performance and computational cost?
\end{enumerate}

\subsection{Extraction vs. Generation for encoder-only PLMs}

To begin with, we investigate the viability of utilizing encoder-only PLMs for KPG by comparing three formulations for present keyphrases: (1) sequence labeling via token-wise classification, (2) sequence labeling with CRF, and (3) sequence generation. The results are presented in Table \ref{tab:main-kpe-results}.

For all the models studied, we find that adding a CRF layer consistently improves the KPE performance. On KP20k, we find that this sequence labeling objective can guide the generation of more accurate (reflected by high F1@M) but much fewer keyphrases, leading to a significantly lower F1@5. On the other hand, the prefix-LM approach allows generating more present keyphrases as well as predicting absent keyphrases, resulting in a much higher F1@5 and a comparable F1@M. On KPTimes, we find sequence labeling achieving better accuracy for BERT and NewsBERT, while prefix-LM achieves a better accuracy for SciBERT.

To summarize, an encoder-only PLM can produce vastly different behaviors depending on the formulation for KPE and KPG. The prefix-LM approach mostly matches the present keyphrase performance of sequence labeling methods and promises to generate many more keyphrases including absent ones. Next, we focus on the two formulations for KPG using encoder-only PLMs and discuss their performance in the context of encoder-decoder PLMs and non-PLM KPG models.

\setlength{\tabcolsep}{3.5pt}
\begin{table*}[]
    \small
    \centering
    \resizebox{0.87\linewidth}{!} {%
    \begin{tabular}{l | c | c c c c | c c c c  }
    \hline
    \multirow{3}{*}{Method} &
    \multirow{3}{*}{|M|} & \multicolumn{4}{c|}{\textcolor{teal}{\textbf{KP20k}}} & \multicolumn{4}{c}{\textcolor{violet}{\textbf{KPTimes}}} \\
    & & \multicolumn{2}{c}{Present} & \multicolumn{2}{c|}{Absent} & \multicolumn{2}{c}{Present} & \multicolumn{2}{c}{Absent}  \\
    & & F1@5 & F1@M & F1@5 & F1@M & F1@5 & F1@M & F1@5 & F1@M \\
    \hline
    \multicolumn{6}{l}{\textit{Encoder-only PLMs}} \\ 
    \hdashline
    BERT-G & 110M & 31.3 & 37.9 & 1.9 & 3.7 & 32.3 & 47.4 & 16.5 & 24.6 \\
    B2B-8+4 & 143M & 32.2 & 38.0 & 2.2 & 4.2 & \underline{33.8} & \underline{48.6} & 16.8 & 24.5 \\
    \textcolor{teal}{SciBERT-G} & \textcolor{teal}{110M} & \textcolor{teal}{\underline{32.8}} & \textcolor{teal}{\underline{\textbf{39.7}}} & \textcolor{teal}{\underline{2.4}} & \textcolor{teal}{\underline{4.6}} & \textcolor{teal}{33.0} & \textcolor{teal}{48.4} & \textcolor{teal}{15.7} & \textcolor{teal}{24.7} \\
    \textcolor{violet}{NewsBERT-G} & \textcolor{violet}{110M} &\textcolor{violet}{29.9} & \textcolor{violet}{36.8} & \textcolor{violet}{1.3} & \textcolor{violet}{2.6} & \textcolor{violet}{33.0} & \textcolor{violet}{48.0} & \textcolor{violet}{17.0} & \textcolor{violet}{\underline{25.6}}  \\   
    \hdashline
    \multicolumn{6}{l}{\textit{Encoder-Decoder PLMs}} \\ 
    \hdashline
    BART & 140M & 32.2 & 38.8 & 2.2 & 4.2 & \textbf{35.9} & \textbf{49.9} & 17.1 & 24.9 \\
    \textcolor{teal}{SciBART} & \textcolor{teal}{124M} & \textcolor{teal}{34.1} & \textcolor{teal}{39.6} & \textcolor{teal}{2.9} & \textcolor{teal}{5.2} & \textcolor{teal}{34.8} & \textcolor{teal}{48.8} & \textcolor{teal}{17.2} & \textcolor{teal}{24.6} \\
    \textcolor{violet}{NewsBART} & \textcolor{violet}{140M} & \textcolor{violet}{32.4} & \textcolor{violet}{38.7} & \textcolor{violet}{2.2} & \textcolor{violet}{4.4} & \textcolor{violet}{35.4} & \textcolor{violet}{49.8} & \textcolor{violet}{17.6} & \textcolor{violet}{\textbf{26.1}}  \\
    \hdashline
    \multicolumn{6}{l}{\textit{Non-PLM baseline}s} \\ 
    \hdashline
    CatSeq & 21M & 29.1 & 36.7 & 1.5 & 3.2 & 29.5 & 45.3 & 15.7 & 22.7 \\
    ExHiRD-h & 22M & 31.1 & 37.4 & 1.6 & 2.5 & 32.1 & 45.2 & 13.4 & 16.5 \\
    Transformer & 98M & 33.3 & 37.6 & 2.2 & 4.6 & 30.2 & 45.3 & 17.1 & 23.1 \\
    SetTrans & 98M & \textbf{35.6} & 39.1 & \textbf{3.5} & \textbf{5.8} & 35.6 & 46.3 & \textbf{19.8} & 21.9 \\
    \hline
    \end{tabular}
    }
    % \vspace{-2mm}
    \caption{
    A comparison across encoder-only and encoder-decoder PLMs from different domains for keyphrase generation. The best results are boldfaced, and the best encoder-only PLM results are underlined. We use different colors for PLMs in the \textcolor{teal}{science} and the \textcolor{violet}{news} domain.
    }
    \label{tab:non_original_seq2seq}
    % \vspace{-2mm}
\end{table*}

\subsection{Can encoder-only PLMs generate better keyphrases than BART and strong non-PLM methods?}
In this section, we compare (1) KPG via prefix-LM, (2) KPG with the best performance of BERT2BERT models with a 12-layer budget (full details in \cref{section-param-allocation}), (3) BART-based seq2seq PLMs, and (4) four strong non-PLM KPG models. Table \ref{tab:non_original_seq2seq} presents the major results on KP20k and KPTimes.

% \begin{figure}[!t]
% \centering
% % \vspace{-6mm}
% \includegraphics[width=0.48\textwidth]{figures/perf-vs-plm-domain.pdf}
% % \vspace{-6mm}
% \caption{Keyphrase generation performance of PLMs pre-trained in different domains. Each table shows the performance of BERT-G, SciBERT-G, and NewsBERT-G in the first row and the performance of BART, SciBART, and NewsBART in the second row.}
% % \vspace{-2mm}
% \label{perf-vs-plm-domain}
% \end{figure}

\begin{figure}[]
\centering
% \vspace{-4mm}
\includegraphics[width=0.48\textwidth]{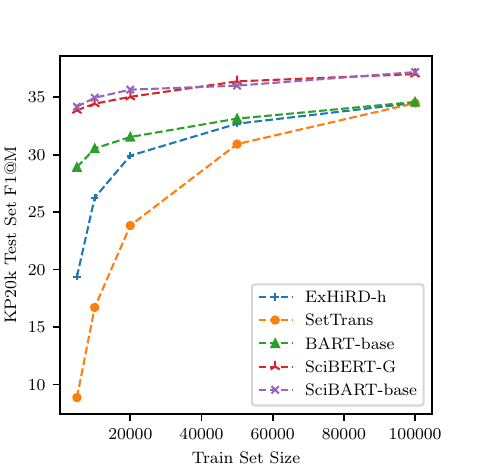}
% \vspace{-6mm}
\caption{Present keyphrase generation performance of different methods as a function of training set size. Fine-tuning in-domain PLMs is much more data-efficient than the other approaches.}
% \vspace{-2mm}
\label{perf-vs-resource-kp20k}
\end{figure}

\paragraph{BERT vs. BART} We start with the surprising result that strong KPG models can be obtained through prefix-LM fine-tuning of encoder-only PLMs. The strength is especially notable with \textit{in-domain} BERT models. On KP20k, SciBERT-G outperforms BART-base on all the metrics. On KPTimes, NewsBERT-G has a comparable F1@5 and better F1@M for absent KPG compared to the BART-base. We also provide the results from in-domain BART models as they have been shown to achieve strong performance gains \citep{wu-etal-2023-rethinking}. We note that such models are often not as accessible as the in-domain BERT model in highly specialized domains like KP20k. Our findings highlight the viability of directly using the in-domain encoder-only PLMs for KPG and obtain comparable results. The trend is more evident in the low-resource scenario. As shown in Figure \ref{perf-vs-resource-kp20k}, SciBERT-G is as resource-efficient as SciBART, outperforming BART in every low-resource setting on KP20k.

% and thus the main emphasis of this section is showing that our approach achieves strong KPG performance with in-domain BERT models 

\paragraph{BERT vs. non-PLM baselines} Given a large training set (500k for KP20k and 250k for KPTimes), the best-performing encoder-only KPG methods outperform SetTrans on KPTimes but only achieve the same or worse performance compared to SetTrans on KP20k. This is expected since the non-PLM architectures introduce task-specific designs to enable more effective training or inference. However, in the low-resource scenario, we find that the preference for encoder-only PLMs is clear. As shown in Figure \ref{perf-vs-resource-kp20k}, SciBERT-G only requires 5k data to achieve the same F1@M of SetTrans fine-tuned with 100k data. 

\paragraph{Prefix-LM vs. BERT2BERT} We observe that combining two smaller-sized BERT models and training on KPG outperform BERT-G despite a similar model size. On KPTimes, the B2B model with an 8-layer encoder and a 4-layer decoder achieves the best present keyphrase performance among all encoder-only PLMs. The model also has \textbf{a lower inference latency} due to its shallow decoder (further explored in \cref{section-param-allocation}). However, this formulation may not be as resource-efficient as prefix-LM, as it randomly initializes cross-attention weights. 

% \textcolor{blue}{ Overall, our results have the important implication that \textbf{in the absence of in-domain seq2seq PLMs, an in-domain encoder-only PLM is preferred over the domain-general BART}. This preference is clearer in low-resource scenarios (Figure \ref{perf-vs-resource-kp20k}): SciBERT only requires 5k data to achieve the same F1@M of BART-base fine-tuned with 100k data.}

To summarize, the two KPG formulations we explore enable encoder-only PLMs to outperform general-domain seq2seq PLMs on KPG, with potentially better data efficiency (when in-domain models are employed) and compute efficiency (when BERT2BERT formulation is used). The results suggest a new possibility of utilizing the more accessible encoder-only models to build stronger keyphrase generation systems, especially in highly specialized domains. 

\subsection{What is the best parameter allocation for BERT2BERT?}
\label{section-param-allocation}

Observing that the BERT2BERT setup produces strong keyphrase generation models, we further investigate the optimal parameter allocation strategy. Specifically, under a given parameter budget, should depth (i.e., more layers) or width (fewer layers, more parameters per layer) be prioritized? Moreover, should the encoder or the decoder be allocated more parameters? As computational resources prevent us from extensive pre-training, we use the BERT2BERT setting with different sized BERT models instead. 

\paragraph{Depth should be prioritized over width.}
We design four pairs of B2B models with different total parameter budgets: 20M, 50M, 85M, and 100M. Each pair contains (a) a model that prioritizes depth and (b) a model that prioritizes width. We ensure that (a) and (b) have similar encoder depth to decoder depth ratios (except the 85M group). The results on KP20k are presented in Table \ref{tab:depth-vs-width}. For all the groups, model (a) performs significantly better despite having slightly fewer parameters. 

\paragraph{A deep encoder with a shallow decoder is preferred.} We fix a budget of 12 layers and experiment with five encoder-decoder combinations. Table \ref{tab:b2b_ablations} presents the results on KP20k and KPTimes. For both datasets, we find that the performance increases sharply and then plateaus as the depth of the encoder increases. With the same budget, \textbf{a deep encoder followed by a shallow decoder is clearly preferred over a shallow encoder followed by a deep decoder}. We hypothesize that comprehending the input article is important and challenging while generating a short string comprising several phrases based on the encoded representation does not necessarily require a PLM-based decoder. 

To verify, we further experiment with randomly initializing either the encoder ("R2B") or the decoder ("B2R"). The results are shown in Table \ref{tab:b2b_ablations}. For both KP20k and KPTimes, we observe that randomly initializing the encoder harms the performance, while randomly initializing the decoder does not significantly impact the performance (the absent KPG is even improved in some cases). 

\setlength{\tabcolsep}{3.5pt}
\begin{table}[!t]
    \small
    \centering
    \resizebox{\linewidth}{!} {%
    \setlength{\tabcolsep}{2pt}
    \begin{tabular}{l | c | c c | c c }
    \hline
    \multirow{2}{*}{Model setup} & \multirow{2}{*}{|M|} & \multicolumn{2}{c|}{\textbf{Present KPs}} & \multicolumn{2}{c}{\textbf{Absent KPs}} \\
    & & F1@5 & F1@M & F1@5 & F1@M \\
    \hline
    B2B-12+12-128 & 13M & \textbf{26.4} & \textbf{33.8} & \textbf{1.0} & \textbf{2.1} \\
    B2B-2+2-256 & 20M & 22.1 & 31.9 & \textbf{1.0} & \textbf{2.1} \\
    \hline
    B2B-12+12-256 & 38M & \textbf{30.8 }& \textbf{36.7} & \textbf{1.6} & \textbf{3.3} \\
    B2B-2+2-512 & 47M & 27.5 & 34.7 & 1.4 & 3.0 \\
    \hline
    B2B-10+4-512 & 81M & \textbf{31.4} & \textbf{37.4} &\textbf{ 1.8 }& \textbf{3.7} \\
    B2B-2+2-768 & 82M & 29.4 & 35.5 &\textbf{1.8} & 3.5 \\
    \hline
    B2B-12+6-512 & 95M & \textbf{31.9} & \textbf{38.2} &\textbf{2.1} & \textbf{4.0} \\
    B2B-4+2-768 & 96M & 30.8 & 37.1 & 2.0 & \textbf{4.0} \\
    \hline
    \end{tabular}
    }
    % \vspace{-2mm}
    \caption{
    Comparison between different parameter allocation strategies. The best performance of each group is boldfaced. B2B-$e$+$d$-$h$ denotes a B2B model with $e$ encoder layers, $d$ encoder layers, and hidden size $h$.
    }
    \label{tab:depth-vs-width}
    % \vspace{-2mm}
\end{table}

\setlength{\tabcolsep}{3.5pt}
\begin{table}[!t]
    \small
    \centering
    % \resizebox{\linewidth}{!}{
    \begin{tabular}{l | c | c | c c | c c  }
    \hline
    \multirow{2}{*}{$e$-$d$} &
    \multirow{2}{*}{|M|} & \multirow{2}{*}{Arch.} &   \multicolumn{2}{c|}{\textbf{KP20k}} & \multicolumn{2}{c}{\textbf{KPTimes}} \\
    & & & F1@5 & F1@M & F1@5 & F1@M \\
    \hline
    \multicolumn{6}{l}{\textbf{Present keyphrase generation}} \\ \hline
    2-10 & 158M & B2B & 30.4 & 36.4 & 31.6 & 46.5 \\ \hdashline
    \multirow{3}{*}{4-8} & \multirow{3}{*}{153M} & B2B & 31.7 & 37.7 & 32.9 & 47.6 \\
     &  & {R2B} & 26.3 & 35.2 & 28.2 & 43.3 \\
     &  & {B2R} & 31.7 & 37.9 & 32.6 & 47.5 \\ \hdashline
    \multirow{3}{*}{6-6} & \multirow{3}{*}{148M} & B2B & 32.1 & 37.7 & \textbf{33.8} & 48.4 \\
     &  & {R2B} & 26.4 & 35.3 & 27.8 & 42.9 \\
     &  & {B2R} & 32.0 & 38.4 & 33.3 & 48.2 \\ \hdashline
     \multirow{3}{*}{8-4} & \multirow{3}{*}{143M} & B2B & \textbf{32.2} & \textbf{38.0} & \textbf{33.8} & \textbf{48.6} \\
    &  & {R2B} & 27.3 & 35.4 & 27.8 & 42.8 \\
     &  & {B2R} & 31.2 & 37.9 & 33.2 & 48.0 \\ \hdashline
    10-2 & 139M & B2B & 31.7 & 38.0 & 33.5 & 48.4 \\
    \hline
    \multicolumn{6}{l}{\textbf{Absent keyphrase generation}} \\ \hline
    2-10 & 158M & B2B & 2.1 & 3.9 & 16.2 & 23.2 \\ \hdashline
    \multirow{3}{*}{4-8} & \multirow{3}{*}{153M} & B2B & \textbf{2.2} & 4.1 & 15.9 & 23.6 \\
    &  & {R2B} & 2.5 & 4.2 & 14.7 & 24.3 \\
    &  & {B2R} & 2.2 & 4.2 & 16.5 & 24.1 \\ \hdashline
    \multirow{3}{*}{6-6} & \multirow{3}{*}{148M} & B2B & \textbf{2.2} & 4.1 & 16.4 & 24.1 \\
    &  & {R2B} & 2.6 & 4.3 & 14.5 & 20.8 \\
    &  & {B2R} & 2.3 & 4.4 & 16.2 & 23.9 \\ \hdashline
    \multirow{3}{*}{8-4 } & \multirow{3}{*}{143M} & B2B & \textbf{2.2} & \textbf{4.2} & \textbf{16.8} & \textbf{24.5} \\
    &  & {R2B} & 2.4 & 4.1 & 14.9 & 21.0 \\
    &  & {B2R} & 2.1 & 4.1 & 16.8 & 24.7 \\ \hdashline
    10-2 & 139M & B2B & 2.1 & 4.1 & \textbf{16.8} & \textbf{24.5} \\
    \hline
    \end{tabular}
    % }
    % \vspace{-2mm}
    \caption{
    A comparison between different BERT2BERT architectures and initializations. In $e$-$d$, $e$ and $d$ indicate the number of encoder and decoder layers, respectively. The best results among the B2B models are boldfaced. 
    }
    \label{tab:b2b_ablations}
    % \vspace{-3mm}
\end{table}

\begin{figure}[t]
\centering
\vspace{-6mm}
\includegraphics[width=0.48\textwidth]{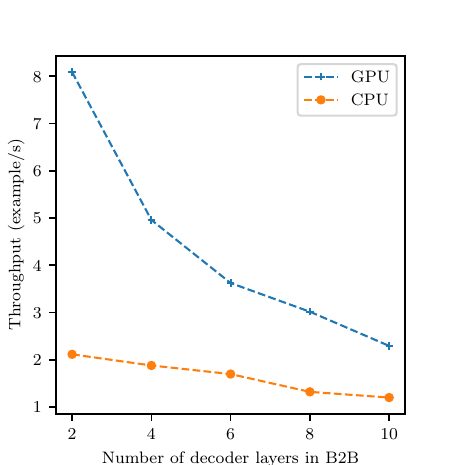}
\caption{Inference speed of BERT2BERT models with different encoder-decoder configurations on GPU and CPU. All the data points are obtained with BERT2BERT models with 12 layers. A model with x decoder layers has 12-x encoder layers.}
\label{b2b-inference-speed}
\end{figure}

\paragraph{Latency Concerns} We further explore the impact on the inference speed of different BERT2BERT configurations. Specifically, we measure and compare the inference throughput of B2B-2+10, B2B-4+8, B2B-6+6, B2B-8+4, and B2B-10+2 in GPU and CPU environments. We use the best model trained on KP20k and test on the KP20k test set with batch size 1, no padding. We use an Nvidia GTX 1080 Ti card for GPU and test on the full KP20k test set. We use a local server with 40 cores for CPU and test on the first 1000 examples from the KP20k test set. No inference acceleration libraries are used. Figure \ref{b2b-inference-speed} reports the averaged throughput (in example/s) across three runs. The throughput decreases significantly with deeper decoders for both CPU and GPU. B2B-8+4 achieves better performance than B2B-6+6 while being 37\% faster on GPU and 11\% faster on CPU.
In conclusion, with a limited parameter budget, we recommend using \textit{more layers} and \textit{a deep-encoder with shallow-decoder} architecture.

% \subsection{Analysis: the domain sensitivity of encoder-only PLMs}

\begin{table}[]
    \small
    \centering
    % \resizebox{\linewidth}{!} {%
    \begin{tabular}{l | c | c | c | c | c }
    \hline
    Model & $D$  & $B_s$ & $E$ & $W_s$ & $LR$ \\
    \hline
    \multicolumn{6}{l}{\textbf{Keyphrase extraction}} \\
    \hdashline
    Transformer & 0.1   & 32 & 10 & 2000 & 3e-5  \\
    BERT  & 0.1   & 32 & 10 & 1000 & 1e-5  \\
    SciBERT  & 0.1   & 32 & 10 & 1000 & 1e-5  \\
    % RoBERTa & 0.1   & 32 & 10 & 1000 & 1e-5  \\
    Transformer+CRF & 0.1   &32 & 10 & 2000 & 3e-5  \\
    BERT+CRF  & 0.1   & 32 & 10 & 2000 & 1e-5  \\
    SciBERT+CRF  & 0.1   & 32 & 10 & 2000 & 1e-5  \\
    RoBERTa+CRF & 0.1   & 32 & 10 & 2000 & 1e-5  \\
    \hline
    \multicolumn{6}{l}{\textbf{Keyphrase generation}} \\
    \hdashline
    BERT-G & 0.1   & 64 & 6 & 4000 & 1e-4 \\
    SciBERT-G & 0.1   & 128 & 6 & 2000 & 1e-4  \\
    % RoBERTa-G & 0.1   & 64 & 6 & 4000 & 1e-4 \\
    UniLM & 0.1   & 128 & 6 & 2000 & 1e-4  \\
    BERT2BERT & 0.0   & 32 & 20 & 2000 & 5e-5  \\
    BART-base & 0.1   & 64 & 15 & 2000 & 6e-5  \\
    SciBART-base & 0.1   & 32 & 10 & 2000 & 3e-5  \\
    % T5-base & 0.1   & 64 & 15 & 2000 & 6e-5  \\
    KeyBART & 0.1   & 64 & 15 & 2000 & 3e-5  \\
    \hline
    \end{tabular}
    % }
    \caption{Hyperparameters for fine-tuning PLMs for keyphrase extraction and keyphrase generation on KP20k. The hyperparameters are determined using the loss on the KP20k validation dataset. We follow a similar set of hyperparameters for KPTimes. $D$ = dropout, $B_s$ = batch size, $E$ = number of epochs, $W_s$ = number of warm-up steps, $LR$ = learning rate. We use early stopping for all the models and use the model with the lowest validation loss as the final model. For all the models, we use weight decay 0.01 and a linear $LR$ schedule.}
    \label{tab:hyperparams}
\end{table}

\section{Conclusion}
This paper unveils the viable application of encoder-only Pre-trained Language Models (PLMs) in the Keyphrase Generation (KPG) task. We introduce two formulations for utilizing BERT-like PLMs for KPG and our systematic experiments demonstrate their strong performance, resource efficiency, and competitive inference latency. Our study opens up promising possibilities for a wider range of in-domain models of efficient and effective keyphrase generation systems.

However, the study is not without limitations. For instance, we only focus on small-sized models below 500M. In addition, the transferability of domain-specific models, as observed between SciBERT and NewsBERT, hints at a potential domain bias that warrants further examination. Future endeavors could delve into a more exhaustive evaluation across a spectrum of encoder-decoder PLMs, investigate the underlying factors affecting model transferability, and explore the augmentation of encoder-only PLMs with additional external knowledge to further enhance keyphrase prediction accuracy across diverse domains. The insights in our work lay a preliminary foundation for more comprehensive inquiries into architecturally diverse PLMs to continually refine KPG methodologies.

\section{Ethics Statement}
RealNews is released under Apache 2.0. We conduct text cleaning and email/URL filtering on RealNews to remove sensitive information. The SciKP and KPTimes benchmarking datasets, distributed by the original authors, are utilized as-is with no additional preprocessing performed before fine-tuning except for lower casing and tokenization. We do not re-distribute any of the pre-training and benchmarking datasets.

The computational resources required for pre-training NewsBART and NewsBERT are significant, and we estimate the total CO$_2$ emissions to be around 600 kg using the \href{https://mlco2.github.io/impact/#compute}{calculation application} provided by \citet{1910.09700}. Additionally, we estimate that all fine-tuning experiments, including hyperparameter optimization, emitted around 1500 kg CO$_2$. To assist the community in reducing the energy required for optimizing PLMs across various NLP applications in the scientific and news domains, we provide the hyperparameter and will release our pre-trained NewsBART and NewsBERT checkpoints. We will limit the access to the models with a non-commercial license. We will also release the raw predictions of our models.

% \section{Acknowledgements}
% The research is partly supported by Taboola and an Amazon AWS credit award. We thank the Taboola team for helpful discussions and feedback. We also thank the anonymous ARR reviewers and the members of the UCLA-NLP group for providing their valuable feedback.

% \newpage
\nocite{*}
\section{Bibliographical References}\label{sec:reference}

% \vspace{-4mm}
\bibliographystyle{lrec-coling2024-natbib}
% \bibliography{lrec-coling2024-example}
\bibliography{custom}

\end{document}